# Can postgraduate translation students identify machine-generated text?

Michael Farrell
IULM University
Milan
Italy
michael.farrell@iulm.it

## Abstract

Given the growing use of generative artificial intelligence as a tool for creating multilingual content and bypassing both machine and *traditional* translation methods, this study explores the ability of linguistically trained individuals to discern machine-generated output from human-written text (HT). After brief training sessions on the textual anomalies typically found in synthetic text (ST), twenty-three postgraduate translation students analysed excerpts of Italian prose and assigned likelihood scores to indicate whether they believed they were human-written or AI-generated (ChatGPT-4o). The results show that, on average, the students struggled to distinguish between HT and ST, with only two participants achieving notable accuracy. Closer analysis revealed that the students often identified the same textual anomalies in both HT and ST, although features such as low burstiness and self-contradiction were more frequently associated with ST. These findings suggest the need for improvements in the preparatory training. Moreover, the study raises questions about the necessity of editing synthetic text to make it sound more human-like and recommends further research to determine whether AI-generated text is already sufficiently *natural-sounding* not to require further refinement.

## 1 Introduction

Authors writing in a second language can today bypass the traditional process of writing in their native language and then having their work translated – either by a human translator or through machine translation – by engineering customized prompts for generative artificial intelligence (GenAI). These prompts, which may be written in the author's native language, the target language or a combination of both, include a precise description, outline or rough draft of the intended text. Consequently, there is no source language document in the traditional sense.

Content generated in this way may then be refined by a human synthetic-text editor tasked with enhancing its engagement and giving it a more human-like tone. This type of editing requires a skill set distinct from that used in post-editing, as the textual anomalies present in synthetic text (ST), – such as redundancy, blandness, verbosity, low burstiness and lack of complex analysis – differ from those typically seen in raw machine translation output (Dou et al. 2022; Farrell, 2025a).

These anomalies appear to be potentially language-independent. For example, redundancy – defined as the repetition of information without adding new meaning or value – can occur in texts written in any language.

The need for synthetic-text editing (STE) assumes that readers are indeed capable of distinguishing AI-generated output from human-written text (HT). Moreover, the ability to identify the textual anomalies characteristic of ST is essential for effective STE.

Clark et al. (2021) observed that untrained, non-expert evaluators are not well equipped to detect machine-generated English text, and even with training, their detection success rate improved only slightly, reaching about 55%. In their study, the evaluators were recruited through Amazon Mechanical Turk and were screened only by location/language (English) and their approval rating on the platform. They did not possess specialized knowledge, such as familiarity with Large Language Models (LLMs) or a background in linguistics[1]. Conversely, Dou et al. found that English ST and HT could be distinguished after laypeople (also recruited through Amazon Mechanical Turk) annotated the texts using a framework called *Scarecrow*, which defines specific error types.

## 2 Aims

The principal objective of this experiment was:

- To evaluate whether postgraduate translation students can effectively identify Italian ST after brief training sessions.

There were also several secondary aims:

- To have the students identify examples of textual anomalies that can be used to enhance the training material, and to determine whether the same categories of ST anomalies found in English texts also occur in Italian.

- To refine the training instructions by identifying areas that require clarification or adjustments to reduce the occurrence of false positives.

- To shed light on the need for STE. If postgraduate translation students cannot reliably distinguish ST, it may already be sufficiently human-like without the need for further editing.

- To assess whether ChatGPT-4o can be guided through prompts to avoid the types of anomalies typically observed in ST.

## 3 Method

Twenty-three postgraduate translation students at the IULM University in Milan, Italy, attended two 30-minute lessons, held one week apart, introducing LLMs, generative artificial intelligence (GenAI) and some common anomalies reported in ST (Dou et al. 2022; Farrell, 2025a). During these lessons, a few examples of textual anomalies were provided, with the hope that the experiment itself would generate additional examples to improve future training materials.

The participants were then presented with 28 short excerpts (ranging from 268 to 467 words) drawn from seven Italian short stories, divided into four sets of seven excerpts each (A, B, C and D). They were informed that each set contained at least one HT and at least one ST excerpt. In reality, each set contained precisely one sequential excerpt of approximately equal length from Alberto Moravia's short story *L'incosciente* (The Reckless Man), from *Racconti romani* (Roman Tales, 1954), along with six sequential excerpts from unabridged short stories generated by ChatGPT-4o using prompts engineered as described below. The order of excerpts was randomized within each set.

The excerpt sets were assigned based on the students' seating arrangement in the lecture room. The student sitting in the first row on the right (from the lecturer's point of view) was assigned set A, the student to their right was assigned set B, the next student set C, and so on, cycling through the sets to ensure a roughly equal distribution. The students were instructed to move on to the next alphabetical set if they finished evaluating their initial set before the allotted time expired. The participants working on set D were instructed to proceed to set A. The experiment concluded once the researcher judged that every student had analysed at least one complete set.

The students were asked to assign a score from 0 to 10 to each text excerpt based on its likelihood of being machine-generated (0 = human-written; 10 = machine-generated; 5 = uncertain). Intermediate integer scores were allowed. They were also asked to identify and classify the types of anomalies or errors that influenced their assessments according to the categories illustrated during the training sessions. Due to time constraints, the participants were encouraged, but not required, to provide specific examples of the anomalies they identified.

To prevent the students from distinguishing the HT excerpts by finding them online, they were not allowed to consult the internet during the experiment. They were also not allowed to speak to other people, including fellow participants.

A few weeks later, a debriefing session was held, where the students were asked to provide feedback on the experiment and training through a preliminary questionnaire, a class discussion and a final questionnaire identical to the first, to determine whether the discussion had caused them to change their opinions.

---

[1] Unpublished clarification courtesy of Elizabeth Clark.

## 3.1 Prompt engineering

A prompt reverse-engineering approach, based on the Automatic Prompt Engineer technique (Zhou et al., 2022), was used because it effectively extracts the storyline from a story, allowing the AI-generated output to follow a similar narrative structure to the human-written one. The aim was to minimize the influence of subjective preferences regarding differing content or theme.

The initial prompt was generated by ChatGPT-4o itself by uploading Alberto Moravia's short story and entering the following instruction:

> "If I had to write a prompt that would cause you to generate the Italian text in the attached file, what would it be? Keep in mind that it is 1808 words long, including the title."

The first artificially generated story (ST1) was then generated by entering the prompt provided by ChatGPT-4o (Appendix A) into a new chat.

The second AI-generated story (ST2) was produced similarly but with modifications to the prompt to set the story in Rome and to name the young protagonists Emilio and Santina, as in Moravia's original. The following additional instruction was also appended to the new prompt:

> "Machine generated text is often criticized for the excessive repetition of words or phrases; the repetition of information without adding new meaning or value; the absence of emotion, creativity and engagement; overly long, highly descriptive, fanciful sentences; uniform sentence structure and length; and lack of complex analysis. Make sure the generated text does not have any of these anomalies."

For ST3, the prompt retained the same setting and character names but replaced the instruction to avoid textual anomalies with:

> "Write the text in the style of the Italian author Alberto Moravia (1907–1990)."

ST4's prompt shifted the setting to a neighbourhood on the outskirts of Naples and the protagonists were renamed Emilio Capuozzo (Mimì) and Santina Picariello (Tina). It also specified that the story should be written in the style of Italian author Elena Ferrante, whose Neapolitan Novels are set similarly.

ST5 was set in Asti, with protagonists Emilio and Santina, and was written in the style of Italian crime writer Giorgio Faletti, a native of Asti.

ST6 moved to Florence, again with Emilio and Santina as protagonists. The requested style was that of the Florentine journalist and author Oriana Fallaci. All six AI-generated stories (ST1–ST6) were produced on 31 August 2024.

In all cases, the prompts specified that the generated Italian short stories should be approximately 1800 words long. However, the AI-generated stories turned out to be shorter than Moravia's original (HT0). To ensure excerpts of comparable length, the last 271 words of HT0 were omitted. Each story was divided into four consecutive excerpts of approximately equal length, avoiding splits mid-paragraph, and one excerpt was placed into each of the four sets of seven (A, B, C and D).

Although ChatGPT-4o was asked to generate similar short stories to reduce the effect of subjective preferences for certain topics, stylistic variation was deliberately introduced by requesting different writing styles based on well-known Italian authors in order to avoid the AI-generated stories being identified due to their similarity.

Lastly, the ST stories and HT0 were analysed using Plagramme AI detector [2] to determine whether any objectively measurable differences existed between them.

## 4 Results

Twenty-three students took part in the experiment. Each one analysed an average of 7.74 text excerpts, with the number of assessments ranging from a minimum of 4 (by 1 participant) and 6 (by 2 participants) to a maximum of all 28 (by 1 participant). As shown in Table 1, on average, the students were unable to identify HT0 since they assigned it an overall mean score of 5.22 (indicating uncertainty). In fact, four of the six AI-generated stories were, on average, perceived as more *human-like* than HT0. None of the short stories were clearly identified as ST, with the highest overall mean score being 5.85 (still very close to uncertain).

However, two students (8.70% of the 23 participants), Student No. 8 and Student No. 20, showed a notable above-average ability to distinguish the HT from the ST excerpts.

Student No. 8 analysed a total of eight excerpts, consisting of all seven excerpts in set C and one excerpt from set A (HT0 A), even though she

---

[2] www.plagramme.com

| Text | Length (words) | Mean Student score | AI Detector score |
|---|---|---|---|
| HT0 A | 359 | 4.13 | 36%[a] |
| HT0 B | 324 | 7.14 | 13% |
| HT0 C | 386 | 5.43 | 16% |
| HT0 D | 467 | 4.00 | 8% |
| **Entire HT0** | **1536**[b] | **5.22**[c] | **17%** |
| ST1 A | 373 | 4.86 | 64% |
| ST1 B | 420 | 6.33 | 100% |
| ST1 C | 408 | 4.71 | 72% |
| ST1 D | 389 | 4.33 | 89% |
| **Entire ST1** | **1590** | **5.04** | **86%** |
| ST2 A | 333 | 1.86 | 97% |
| ST2 B | 392 | 2.17 | 81% |
| ST2 C | 338 | 4.86 | 94% |
| ST2 D | 268 | 3.00 | 100% |
| **Entire ST2** | **1331** | **3.00** | **83%** |
| ST3 A | 375 | 1.67 | 94% |
| ST3 B | 398 | 3.67 | 86% |
| ST3 C | 399 | 3.43 | 87% |
| ST3 D | 376 | 5.67 | 89% |
| **Entire ST3** | **1548** | **3.60** | **85%** |
| ST4 A | 373 | 6.43 | 65% |
| ST4 B | 334 | 7.60 | 96% |
| ST4 C | 367 | 4.57 | 94% |
| ST4 D | 374 | 2.60 | 87% |
| **Entire ST4** | **1448** | **5.33** | **83%** |
| ST5 A | 420 | 2.00 | 94% |
| ST5 B | 379 | 4.00 | 93% |
| ST5 C | 417 | 4.14 | 86% |
| ST5 D | 410 | 3.40 | 82% |
| **Entire ST5** | **1626** | **3.36** | **94%** |
| ST6 A | 306 | 7.29 | 84% |
| ST6 B | 341 | 6.60 | 99% |
| ST6 C | 339 | 5.29 | 89% |
| ST6 D | 331 | 4.43 | 93% |
| **Entire ST6** | **1317** | **5.85** | **73%** |

Table 1: Average scores assigned to the text excerpts by the students.

a) This excerpt includes a paragraph that received an anomalous score of 99% according to Plagramme AI detector.

b) To keep the excerpts to approximately the same length, the last 271 words were not used.

c) The overall mean in each case does not equal the mean of the partial means because the number of students evaluating each excerpt varies.

should have moved on to set D. She accurately assigned a score of 0 to both excerpts written by Alberto Moravia. For the remaining six ST excerpts, she gave scores ranging from 5 (uncertain) to 10 (definitely machine-generated).

Student No. 20, on the other hand, analysed the seven excerpts in set D. The text to which she gave the lowest score (3) – signifying it appeared the most human – was the only HT excerpt she assessed. She assigned relatively high scores, ranging from 7 to 9 (indicating somewhere between probably and almost definitely AI-generated) to the six ST excerpts.

If we exclude the excerpt from set A that Student No. 8 analysed, the two students become directly comparable, since they both evaluated a complete set of seven excerpts, each containing one HT.

Now, let's calculate the probability that these two students correctly identified the HT excerpt purely by chance. The participants were told that at least one excerpt in each set was human-written and at least one was AI-generated. Based on this, there are six possible scenarios per set, ranging from "only one excerpt is HT" to "six of the seven excerpts are HT". Hence, the probability of a student guessing that only one of the seven excerpts is written by a human is 1/6.

Assuming they correctly guess that there is only one HT excerpt, the probability of guessing which one it is without looking at them is 1/7, as each set contains seven excerpts. Since these two guesses are independent, the combined probability of making both guesses correctly is 1/6 * 1/7 = 1/42, or approximately 2.38%.

However, 23 students took part in the experiment, and two of them identified the HT excerpt. The probability that at least two participants out of 23 guess correctly without analysing the excerpts can be calculated using the binomial probability formula:

$$P(k) = \binom{n}{k} p^k (1-p)^{n-k}$$

Where n represents the total number of participants (n=23), k is the number of successful students (k=2), and p is the probability that an individual participant guesses correctly (p=1/42), the probability that at least two students out of 23

| Anomaly | Text | | | | | | |
|---|---|---|---|---|---|---|---|
| | H0 | S1 | S2 | S3 | S4 | S5 | S6 |
| Excessive repetition of words or phrases | **13** | 10 | 4 | 8 | 4 | 5 | 11 |
| Redundancy | **9** | 8 | 3 | 7 | 6 | 5 | 8 |
| Non-existent words | 1 | 0 | 0 | 0 | 0 | **1** | 0 |
| Blandness | **10** | 7 | 6 | 3 | 7 | 3 | 6 |
| Verbosity | **11** | 2 | 3 | 1 | 3 | 3 | 2 |
| Low burstiness | 4 | 9 | 5 | 9 | **9** | 8 | 8 |
| Lack of complex analysis | 7 | **10** | 7 | 5 | 6 | 3 | 5 |
| Grammar and spelling mistakes | **11** | 6 | 7 | 2 | 6 | 4 | 9 |
| Hallucination | **4** | 3 | 0 | 1 | 3 | 0 | 2 |
| Self-contradiction | 2 | 1 | 1 | 1 | **3** | 2 | 2 |
| Unnecessary technical jargon | **4** | 0 | 0 | 0 | 0 | 2 | 0 |
| *Total replies from the students** | *27* | *26* | *25* | *25* | *24* | *25* | *26* |

Table 2: Textual anomalies detected by the students by text. The highest scores are highlighted in bold red.
*This number exceeds the total number of participants because some students analysed more than one excerpt from the same short story.

succeed purely by chance is approximately 10.32%[3].

This relatively low probability strongly suggests that the two students in question used analytical skills, rather than random guessing, to distinguish the AI-generated excerpts from the human-written ones during the experiment.

### 4.1 Textual anomalies detected

Table 2 clearly shows that the participants found most of the textual anomalies they were asked to detect in both the HT and ST excerpts. In fact, the human-written story was perceived as the most artificial text in 7 out of the 11 categories.

Despite this, the results support the assumption that the same ST anomaly categories observed in English ST also occur in Italian ST, with the possible exception of non-existent words, which were absent (see Section 6.2.1), and the notable exception of grammar and spelling mistakes. While such mistakes are relatively rare in artificially generated English texts (Dou et al., 2022; Gillham, 2024), they were found to be common in the Italian ST excerpts (see Section 6.2.3).

Owing to the time constraints mentioned earlier, not all the students provided specific examples of the anomalies they reported: 19 gave examples of grammar and spelling mistakes, 16 of excessive repetition of words or phrases, 11 of redundancy, 9 of low burstiness, 7 of verbosity, 7 of hallucination, 6 of self-contradiction, 6 of unnecessary technical jargon, 3 of lack of complex analysis, 2 of non-existent words (both spurious) and 2 of blandness.

The examples of burstiness and self-contradiction may be used in the future to enhance the training material (see Section 7).

### 4.2 Debriefing

Only eight students attended the debriefing session held a few weeks after the experiment. A ninth student joined later, but her replies were not analysed because she had not completed the initial questionnaire.

During the session, the students were shown the overall mean scores in Table 1 and asked to complete a closed-answer questionnaire on why so many of them had failed to distinguish between the ST and HT excerpts. This was followed by an open discussion covering the questionnaire topics, the experiment itself, the preparatory training and general observations about ST.

After the discussion, the participants were asked to complete the same questionnaire again, with exactly the same questions, to determine whether the classroom discussion had altered their opinions.

Half of the participants, including the only successful student present, stated that the textual anomalies they were asked to look out for could

---
[3] Using the Statology binomial distribution calculator: www.statology.org/binomial-distribution-calculator

Accepted for MT Summit 2025, Geneva, Switzerland, 23-27 June 2025

also be found in HT0. Despite this, at the beginning of the session, three-quarters of the participants, including the successful student, disagreed with the hypothesis that searching for textual anomalies is an ineffective method for identifying ST. However, following the discussion, this proportion dropped to just over one-third (37.5%), although the successful student maintained her original position.

None of the students found the text excerpts too long, and the majority after discussion (62.5%) did not feel they needed to be longer.

Possibly due to a growing sense of disappointment, the percentage of the students who disagreed with the hypothesis that humans cannot distinguish between ST and HT, regardless of the training received, dropped from 75% at the beginning of the classroom discussion to 37.5% by the end. However, no one explicitly agreed with the proposition.

Following the classroom discussion, half of the students deemed the training insufficient and expressed the need for more practice.

The results for the most significant questions, both before and after the discussion, are shown in Appendix B (Table 3 to Table 8). The replies of the only successful student present, Student No. 20, are highlighted in bold red.

## 5 Limitations

Since this experiment was conducted as part of a postgraduate degree course with set number of hours, it was necessarily limited to a small selection of texts of a similar kind in a single language. The time available for preparatory training in GenAI detection was also limited. Moreover, the size of the class restricted the number of participants. As a result, the findings and conclusions of this study may not be broadly generalizable. However, the practical, hands-on learning experience and potential contribution to course development outweigh these limitations.

## 6 Discussion

### 6.1 Postgraduate translation students as evaluators

Judging from the results shown in Table 1 and Table 2, the answer to the question posed in the title of this paper (*Can postgraduate translation students identify machine-generated text?*) appears, at first glance, to be a resounding no.

This result is all the more disappointing considering that postgraduate translation students possess a background in linguistics, which might make them more qualified than the evaluators used in the two studies mentioned in the introduction (Clark et al., 2021; Dou et al., 2022).

In contrast, Plagramme AI detector showed little doubt in its assessments, assigning the ST stories probabilities of being AI-generated of between 73% and 94%, while attributing only a 17% likelihood to Alberto Moravia's work. Moreover, there was no alignment between the AI detector's scores and the mean scores given by the students.

However, closer analysis of individual participant data, as noted in Section 4, reveals that two out of the 23 students involved in the experiment are very probably able to distinguish ST from HT, at least as regards the specific texts analysed in this study.

### 6.2 Textual anomalies

As mentioned in Section 4.1, the students identified most of the textual anomalies they were asked to look out for in both the HT and ST excerpts. This finding was further confirmed during the debriefing session, as mentioned in Section 4.2. The following subsections provide a more detailed discussion of the results regarding specific anomalies.

### 6.2.1 Non-existent words

The student who reported non-existent words in the ST excerpts clarified in a note that she was not actually identifying non-existent words but rather pointing out the unusual use of certain terms. Similarly, the student who flagged a non-existent word in HT0 explained that she was referring to the French word *parabrise*, which – though uncommon – is occasionally used in Italian prose[4]. Neither of these cases involves truly non-existent words, like the term *grasitating* reported in an earlier experiment by Farrell (2025a).

It should be noted in fairness that the participants were not allowed to use a browser to ensure they could not identify which story was human-written by finding parts of it online. Consequently, they were unable to verify the existence of any unusual terms they encountered.

---

[4] www.treccani.it/enciclopedia/ricerca/parabrise/?search=parabrise

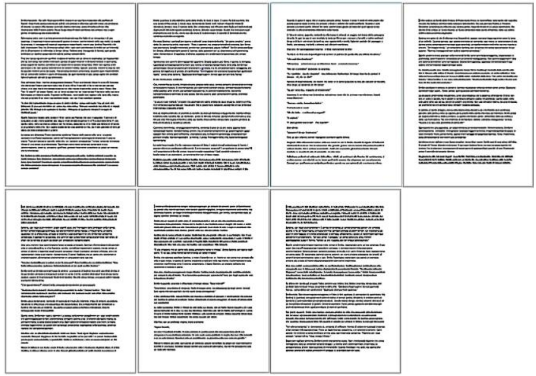

Figure 1: Thumbnails of the first page of each story in print layout view.

In any case, non-existent words could theoretically occur in HT as a result of typos, potentially leading to false positives (see Section 6.2.3).

### 6.2.2 Low burstiness

Burstiness measures variation in writing patterns, including sentence structure and length. Unlike machines, humans tend to exhibit high burstiness by naturally varying their writing to prevent repetition, such as by avoiding multiple sentences that start in the same way. Table 2 shows that most of the students who reported low burstiness correctly associated it with the ST excerpts. When Student No. 8 was asked how she had been so successful, she explained that, in her opinion, the key to identifying ST was noting the use of simple, very short sentences, adding that this brevity was clearly not intended for stylistic purposes.

These findings suggest that low burstiness was the most effective indicator in this experiment among the anomaly categories analysed. Notably, burstiness is also one of the parameters measured by AI detectors, such as GPTZero (Chaka, 2023).

Interestingly, it appears that the burstiness of the stories used in this specific experiment can be roughly estimated simply by examining their print layout, provided there is enough text. To test this idea, a small additional experiment was conducted with six randomly chosen undergraduate translation students from the same university. They were shown illegible thumbnails of the first page of the seven short stories used in the postgraduate experiment, presented in random order, and asked whether any of them stood out in terms of layout.

All six students unequivocally indicated HT0 (the third from the left in the top row of Figure 1). It probably stands out due to its greater use of dialogue, which is also found to a lesser extent in the six ST stories. It would be useful to investigate whether this quick, simple detection method can be generalized to other texts, authors, genres and languages.

### 6.2.3 Grammar and spelling mistakes

Grammar and spelling mistakes are known to be relatively rare in artificially generated English texts (see Section 4.1). However, they are more common in AI-generated Italian texts. In this experiment, the students identified a few examples, including:

1. *Nei giorni seguenti, Emilio evitò Santina, temendo che lei potesse capire cosa **stava** succedendo.*[5]
   Correct Italian grammar requires the use of the subjunctive tense "…**stesse** succedendo".
2. *Santina lo fissò, sorpresa, ma Emilio non cercò **il suo** approvazione.*[6]
   The article and possessive adjective should agree with the noun "…**la sua** approvazione".

The students also identified grammar issues in HT0. However, it is likely that Moravia intentionally used unconventional grammar, such as *"per me, io ci sto"*.[7] as a stylistic device to reflect the social and cultural backgrounds of the characters in his stories, thereby adding authenticity. Indeed, he wrote the story used in this experiment *The Reckless Man* in the first person, imagining himself as a young working-class boy in post-war Rome. Moreover, typos are not uncommon in printed texts, meaning that an error like Example 2 above could also theoretically appear in HT.

Given these factors, grammatical accuracy and spelling seem to be highly unreliable parameters for distinguishing between Italian ST and HT.

### 6.2.4 Hallucination and self-contradiction

All but one of the instances of hallucination reported in HT0 were, in reality, unusual or antiquated turns of phrase (for instance, *non posi*

---

[5] Over the next few days, Emilio avoided Santina, fearing that she might realize what was going on.
[6] Santina stared at him in surprise, but Emilio didn't seek her approval.
[7] Count me in.

*tempo in mezzo*[8]). If the students had been allowed to consult the internet, they would probably have discovered that these expressions exist and might not have flagged them as hallucinations. The remaining example was *custode del passaggio a livello*. While referring to level-crossing guards may seem *hallucinatory* today, they did exist in Italy at the time when Moravia's story was set.

All the cases of hallucination reported in the ST excerpts could just as easily be classified as self-contradictions. Given this, it seems advisable to avoid using the term *hallucination* and instead ask evaluators to focus on identifying self-contradiction. Notably, on her task feedback form, successful Student No. 20 observed that HT0 was the only text to mention specific places in a consistent way.

### 6.2.5 Unnecessary technical jargon

The four students who noted unnecessary technical jargon in HT0 all cited the same two examples: *grassazione*[9] and *rettifilo*[10]. These uncommon terms appear to be part of Alberto Moravia's idiolect, suggesting that this category is prone to producing false positives. The unreliability of this criterion for determining artificial-generated Italian text is one of the key findings of this study.

### 6.2.6 Other anomaly categories

According to the data in Table 2, none of the remaining categories proved effective in helping participants identify the ST excerpts.

### 6.3 Preparatory training

Since there was no initial control experiment conducted without preparatory training, it is hard to determine whether the training contributed to the success of the two students who performed well. Regardless, the fact that only 2 out of the 23 students (8.70%) were able to identify ST after training cannot be considered a successful outcome. Furthermore, as reported in Section 4.2, following the classroom discussion, half of the students deemed the training insufficient.

Successful Student No. 8 mentioned that, over the past year, she had often used GenAI tools (particularly ChatGPT) for reformulating, summarizing and occasionally translating texts, which are among the tasks some professional translators report they use GenAI for in their workflow (Farrell, 2025b). She suggested that this experience had helped her become familiar with the "distinctive writing style of GenAI". Taken together, these observations highlight the importance of providing training on how to use GenAI effectively for such tasks in translation courses.

### 6.4 Text excerpt length

Clark et al. (2021) truncated their text excerpts at the first end-of-sentence after reaching 100 words, while Dou et al. (2022) used whole paragraphs ranging from 80 to 145 tokens. In contrast, this experiment used sequential excerpts of between 268 and 467 words (Table 1). As mentioned in Section 4.2, none of the students found the excerpts too long, and after the discussion, the majority did not feel they needed to be any longer.

However, it seems plausible that low burstiness and self-contradiction would be easier to identify in longer excerpts (see also Section 6.2.2). In the case of short stories, these excerpts could potentially consist of the entire text.

### 6.5 Prompting to avoid anomalies

ST2 was generated using a prompt that specifically instructed ChatGPT-4o to avoid most of the tell-tale textual anomalies the students had been trained to identify. This seems to have been effective, since this story was rated, on average, as the most human-like of the seven analysed, with an overall mean score of 3.0. However, this prompting did not seem to successfully mislead Plagramme AI detector, even though ST2 received the second-lowest probability of being artificial among the six AI-generated stories (83%).

### 6.6 Need for synthetic-text editing

The need for translation students to be familiar with STE techniques stems from the hypothesis that demand for *traditional translation* is likely to decline, while demand for STE will probably grow.

The existence of this demand is in turn based on two assumptions: first, that readers are able to distinguish between ST and HT, and second, that they actually prefer reading HT.

Regarding the first assumption, as noted in Section 4.2, none of the students in this study considered distinguishing between ST and HT a pointless exercise, and the findings suggest that some individuals are indeed capable of doing so.

---

[8] Old-fashioned way of saying "I didn't stop for a moment".
[9] Armed robbery.
[10] Straight stretch of road.

As for the second assumption, a study by Zhang and Gosline (2023) found that advertising content (in English) generated by GenAI, as well as human-created advertising content augmented by GenAI (i.e., automatically edited), was perceived as higher quality than content produced solely by human experts. Similarly, a study by Porter and Machery (2024) revealed that AI-generated poetry is indistinguishable from human-written verse and is rated more favourably.

Consequently, it would seem that STE may not be necessary for all genres of text.

## 7 Conclusion

The low number of students who were able to distinguish between the two kinds of text in this experiment, even after training, suggests that the guidance given needs redesigning. The examples highlighted by the participants indicate that the preparatory exercises should focus on identifying self-contradiction and assessing variability in syntactic structures and lexical distributions, known as burstiness.

Moreover, general training on the use of GenAI as a tool in the translation process, apart from being an essential part of any modern translation course, could also help students better identify ST.

Regarding the length of texts, it would be worthwhile experimenting with longer excerpts.

Lastly, further research should also explore whether readers genuinely prefer human-written text and whether STE, which seeks to make ST sound more human-like, is actually necessary at all.

## Carbon impact statement

The study described in this paper involved seven queries made using ChatGPT-4o. According to a widely cited figure (Wong, 2024), each query generates approximately 4.32 g of $CO_2$ emissions. As a result, the entire experiment produced an estimated total of 30.24 g of $CO_2$, excluding the emissions generated from several hours of internet browsing for background research.

## Acknowledgments

The research project reported in this paper has received funding from the International Center for Research on Collaborative Translation at the IULM University, Milan, Italy.

*Content Generation.* Cambridge University Press. https://doi.org/10.1017/jdm.2023.37

Yongchao Zhou, Andrei Ioan Muresanu, Ziwen Han, Keiran Paster, Silviu Pitis, Harris Chan, Jimmy Ba. 2022. *Large language models are human-level prompt engineers.* Published as a conference paper at ICLR 2023. https://doi.org/10.48550/arXiv.2211.01910

## Appendix A

The initial prompt generated by ChatGPT-4o was:

*Generate a text in Italian that is approximately 1800 words long, including the title. The text should be a short story that explores themes of fear, courage, and moral dilemmas. It should feature a young protagonist who, after being influenced by a romantic interest, decides to write a threatening letter to the owner of a villa. The story should include vivid descriptions of the setting, the protagonist's thought process, the actual writing and delivery of the letter, and the psychological consequences that follow. The narrative should convey the protagonist's initial bravado, followed by increasing anxiety and fear as the reality of their actions sets in. The story should conclude with the protagonist retrieving the letter in a desperate attempt to avoid the consequences of their actions, only to be left questioning their courage and moral standing.*

## Appendix B

Replies to the debriefing questionnaire before and after the class discussion.

|  | Before | After |
| --- | --- | --- |
| I agree | 0 | 0 |
| Maybe | 1 | 4 |
| I disagree | **6** | **3** |
| I don't know | 1 | 1 |

Table 3: The task is pointless. Humans cannot distinguish between ST and HT, regardless of the training they receive.

|  | Before | After |
| --- | --- | --- |
| I agree | 4 | 4 |
| Maybe | **3** | **1** |
| I disagree | 0 | 0 |
| I don't know | 1 | 3 |

Table 4: The preparatory training was insufficient. Additional practice is needed to effectively identify textual anomalies.

|  | Before | After |
| --- | --- | --- |
| I agree | 0 | **4** |
| Maybe | **2** | 4 |
| I disagree | 4 | 0 |
| I don't know | 2 | 0 |

Table 5: Some of the textual anomalies we were searching for can also be found in HTs.

|  | Before | After |
| --- | --- | --- |
| I agree | 0 | 1 |
| Maybe | 1 | 2 |
| I disagree | **6** | **3** |
| I don't know | 1 | 2 |

Table 6: Searching for textual anomalies is not the right approach for this task. The results would have been better if we had relied on intuition.

|  | Before | After |
| --- | --- | --- |
| I agree | 1 | 0 |
| Maybe | 3 | 0 |
| I disagree | 3 | **7** |
| I don't know | **1** | 1 |

Table 7: The texts were too lengthy. The task would have been easier if shorter excerpts had been provided.

|  | Before | After |
| --- | --- | --- |
| I agree | 0 | **2** |
| Maybe | **1** | 1 |
| I disagree | 6 | 5 |
| I don't know | 1 | 0 |

Table 8: The texts were too brief. The task would have been easier if we had been given the entire short story.